\documentclass{article} 
\usepackage{iclr2016_conference,times}
\usepackage{hyperref,algorithm,algpseudocode,array,tabularx,multirow,caption,subcaption,amsfonts,url,verbatim,enumitem,amsmath,graphicx}
\usepackage{url}

\title{Fast Parallel SAME Gibbs Sampling on \\ General Discrete Bayesian Networks}

\author{Daniel Seita, Haoyu Chen \& John Canny \\
Computer Science Division \\
University of California, Berkeley \\
Berkeley, CA 94720, USA \\
\texttt{\{seita,haoyuchen,canny\}@berkeley.edu}
}

\begin{document}

\maketitle

\begin{abstract}
A fundamental task in machine learning and related fields is to
perform inference on Bayesian networks. Since exact inference takes
exponential time in general, a variety of approximate methods are
used.  Gibbs sampling is one of the most accurate approaches and
provides unbiased samples from the posterior but it has historically
been too expensive for large models. In this paper, we present an
optimized, parallel Gibbs sampler augmented with state replication (SAME or State
Augmented Marginal Estimation) to decrease convergence time. We find
that SAME can improve the quality of parameter estimates while accelerating
convergence.  Experiments on both synthetic and real data
show that our Gibbs sampler is substantially faster than the state of
the art sampler, JAGS, without sacrificing accuracy. Our ultimate
objective is to introduce the Gibbs sampler to researchers in many
fields to expand their range of feasible inference problems.
\end{abstract}

\section{Introduction}\label{sec:intro}

In many machine learning applications, the user has a distribution
$P(X,Z \mid \Theta)$ where $X$ is observed data, $Z$ is hidden
(latent) data, and $\Theta$ represents the model parameters. The goal
is generally to find an optimal $\Theta$ with respect to $X$, while
marginalizing out $Z$. To represent these problems, it is common to
use graphical models.  Here we focus on discrete-state models,
so node transitions are characterized by a set of conditional
probability tables (CPTs). Each CPT represents a local probability
distribution $\Pr(X_i \mid X_{\pi_i})$ where $X_i$ is a random
variable, and $X_{\pi_i}$ represents its set of parent nodes in the
graph. We denote the full set of CPTs as $\Theta$.  In general, the
network state is partially observed $\mathcal{D} = \{\xi_1, \ldots,
\xi_m\}$, where $\xi_i$ is an $n$-dimensional vector with assignments
to the $n$ variables of the graph, or ``N/A'' to indicate missing
data. We assume that the structure of the Bayesian network --- its
nodes and edges --- is known in advance, but not the set of
observations. This form allows us to represent latent and well as
observed states, but also observations on an arbitrary
(sample-dependent) subset of those states.

Well-known strategies for parameter estimation with partially observed
data include Expectation-Maximization~\citep{EMpaper} and variations
of gradient ascent~\citep{Thiesson95}.  Parameter estimation using
these methods marginalizes over missing states, which typically
dominates the runtime. Using Monte-Carlo methods for marginalization
leads to the MC-EM methods~\citep{MC-EM}.

For Bayesian parameter estimation with a Gibbs sampler, we sample both latent states and parameters
from $P(Z,\Theta \mid \mathcal{D})$.  For the case of discrete node states, the $Z$ follow
multinomial distributions while the $\Theta$ are Dirichlet. Gibbs sampling is widely used in machine
learning but has seen limited use on large datasets.  It is typically orders of magnitude slower
than special-purpose, approximate methods; for Latent Dirichlet Allocation (LDA), these methods
include Variational Bayes~\citep{Blei2003} and Walker's alias method~\citep{WalkersAlias_2014}.  In
a recent result,~\citet{SAME2015} showed that by combining the State Augmented Monte Carlo (SAME)
technique~\citep{SAME2002} with Gibbs sampling, one can match the speed of approximate methods while
obtaining higher quality estimates. But that paper was restricted to LDA and similar models.  In
this paper, we build upon that result by presenting a SAME Gibbs sampler for general discrete
Bayesian networks. The contributions of this paper are:

\begin{itemize}[noitemsep]
    \item A more general Gibbs sampler using SAME sampling with improved convergence and better-quality MAP or ML parameter
      estimates over standard samplers.
    \item We parallelize the sampler which provides acceleration on both CPU and GPU
      hardware.
    \item The sampler maintains only model-related state and processes data out-of-memory
      so it can scale to very large datasets (either from disk or network storage).
\end{itemize}

We benchmark our sampler versus a state of the art Gibbs sampler,
JAGS~\citep{JAGS2003}, and find that our Gibbs sampler is at least
an order of magnitude faster. The throughput and scalability of the
sampler (nodes processed per second) is competitive with special
purpose methods while maintaining the accuracy of Gibbs sampling.

\section{Related Work}\label{sec:related_work}

The problem of Bayesian inference for graphical models is  well-studied (\citet{Koller2009}
and~\citet{Wainwright2008} cover recent approximate and MCMC methods). Gibbs sampling has proved to
be one of the most important practical techniques for large models; collapsed samplers are widely
used for LDA and related models~\citep{Griffiths_Steyvers}. In addition, parallelism has been
improved using color groups~\citep{Gonzalez2011}, and approximate, uncoordinated parallelism has
been shown to give good results in practice~\citep{Johnson2013}.  Gibbs sampling has also been
applied to large-scale databases~\citep{Zhang2013}.

Recently, Graphics Processing Units (GPUs) have been valuable for deep learning, and are used in
most toolkits including Theano \citep{Theano2012}, CAFFE ~\citep{jia2014caffe} and Torch
\citep{Torch}. We continue this trend by using GPU acceleration for our sampler.

The result most directly related to our paper, as briefly mentioned in Section~\ref{sec:intro}, is
one that shows how the addition of SAME to a GPU-accelerated Gibbs sampler can be very fast for
Latent Dirichlet Allocation and the Chinese Restaurant Process~\citep{SAME2015}. In that paper, they
explored the application of SAME to graphical model inference on modern hardware, and showed that
combining SAME with factored sample representation (or approximation) gives throughput competitive
with the fastest symbolic methods, but with potentially better quality. We extend that result by
implementing a general-purpose Gibbs sampler that can be applied to arbitrary discrete graphical
models.

\section{Fast Parallel SAME Gibbs Sampling}\label{sec:same}

SAME is a variant of MCMC where one artificially replicates latent states to create distributions that
concentrate themselves on the global modes~\citep{SAME2002}. It is an efficient way of performing
MAP estimation in high-dimensional spaces when needing to integrate out a large number of variables.
Given a distribution $P(X,Z\mid \Theta)$, to estimate the most likely $\Theta$ based on the data
$(X,Z)$ using SAME, one would define a new joint $Q$:
\begin{equation}\label{eq:same}
Q(X,\Theta,Z^{(1)},\ldots,Z^{(m)}) = \prod_{j=1}^m P(X,\Theta,Z^{(j)})
\end{equation}
which models $m$ copies of the distribution tied to the same set of parameters $\Theta$, which in
our case forms the set of Bayesian network CPTs. This new distribution $Q$ is proportional to a
\emph{power} of the original distribution, so $Q(\Theta \mid X) \propto (P(\Theta \mid X))^m$. Thus,
it has the same optima, including the global optimum, but its peaks are sharpened~\citep{SAME2002}.
Note that as $m$ increases, SAME approaches Expectation-Maximization~\citep{EMpaper} since the
distribution would peak at the value corresponding to the maximum likelihood estimate.

We argue that SAME is beneficial for Gibbs sampling because it helps
to reduce excess variance derived from discrete sample quantization.
It is important, however, not to set the SAME replication factor $m$
too high, which reduces sampling to EM and may lead to poor local
optima. Instead, since the replication factor is formally equivalent
to a temperature control, it can be used to anneal the sampler to a
low-variance, high-quality parameter estimate.


\section{Implementation of SAME Gibbs Sampling}\label{sec:implementation}

\begin{figure}[t]
\centering
\includegraphics[width=0.75\textwidth]{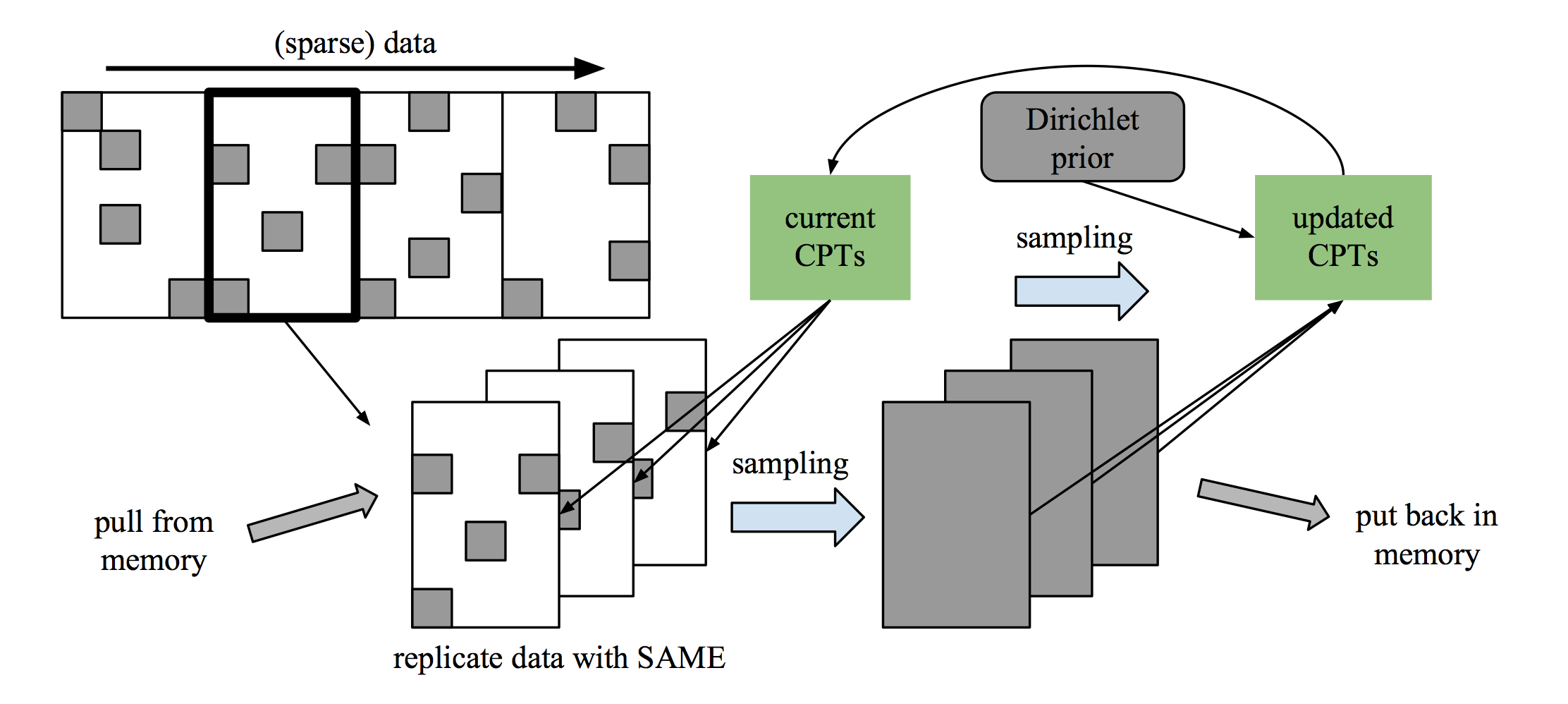}
\caption{This visualizes our Gibbs sampler at work. The original data is split into four
minibatches of equal sizes (for caching purposes), with each having some known data (shaded gray)
and unknown data (white). For each minibatch, the sampler replicates the data three times ($m=3$),
samples the unknown values, and then uses those with the prior to update its estimate of the CPTs.}
\label{fig:BIDMach}
\end{figure}

Our Gibbs sampler is implemented as part of the open-source BIDMach library~\citep{bidmach} for
machine learning.  Figure~\ref{fig:BIDMach} shows a visualization of how it works on a large dataset.
Our sampler expects a data matrix (typically sparse), with rows representing variables and columns
representing cases. BIDMach divides data into same-sized \emph{minibatches} and iterates through
them to update parameters. If there are $M$ minibatches in the entire dataset, we maintain a moving
sum of the state counts for the $M$ most-recently processed minibatches. Updating this average requires 
us to maintain the state counts for each minibatch from the last iteration. We subtract these from
the total state counts before adding in the updated counts for this minibatch from the current
iteration\footnote{For larger datasets, this state is not saved and instead the counts are
maintained using an exponential moving sum. Thus the memory require scales only with model size, not
dataset size.}.

Our Gibbs sampler is augmented with SAME. Consequently, if $m$ is the SAME parameter, for each
minibatch our sampler forms $m$ copies of the known data.  Then, it performs Gibbs sampling to fill
in the missing data in each copy of the minibatch using the current CPTs.  These sampled results
are combined with an adjustable Dirichlet prior and the current CPTs to form a set of discrete
counts, which are the Dirichlet parameters to update (via sampling) the CPTs. 

There are several optimizations used to improve performance.  First, since storage allocation is
very expensive on GPUs and their memory is limited (3-12 GB is typical), BIDMach uses a matrix
caching strategy to reuse memory for matrices of the same dimensions. (This is why minibatches in
BIDMach need to be the same size.) Second, graphical structure and SAME-replicated states for a
particular color group are fused into large matrices to maximize parallelism and hide GPU-kernel
overhead. The sampling process itself is done via a series of matrix multiplication operations.

As mentioned in Section~\ref{sec:related_work}, we further parallelize Gibbs sampling in an
exact manner via chromatic partitioning. In Bayesian networks, nodes $u$ and $v$ are independent
conditioned on a set of variables $\mathcal{C}$ if $\mathcal{C}$ includes at least one variable on
every path connecting $u$ and $v$ in the \emph{moralized graph} of the network, which is the graph
formed by connecting parents and dropping edge orientations.

Suppose there is a $k$-coloring of the moralized graph. Let $\mathcal{V}_c$ be the set of variables
assigned color $c$ where $1 \leq c \leq k$. One can sample sequentially from $\mathcal{V}_1$ to
$\mathcal{V}_k$, and within each color group, sample all its variables in parallel. This parallel
sampler corresponds exactly to the execution of a sequential scan Gibbs sampler for some permutation
over the variables and will converge to the desired distribution because variables within one color
group are independent given all the other variables. Finding the optimal coloring of a general graph
is NP-complete, so we use efficient heuristics~\citep{kubale2004graph} for balanced graph coloring,
which work well in practice.

\section{Evaluating our Gibbs Sampler}\label{sec:experiments}

We benchmark our Gibbs sampler based on one synthetic and one real dataset. We compare it with
JAGS~\citep{JAGS2003}, which is the most popular and efficient tool for Bayesian inference, and also
uses Gibbs sampling as the primary inference algorithm. 


For all JAGS experiments and for CPU benchmarks for BIDMach, we use a
single computer with an Intel Core Xeon processor with eight cores and
2.20 GHz clock speed (E5-2650 Sandy Bridge). The computer has 64 GB of
CPU RAM. We used a different machine with an NVIDIA Titan X GPU for the
GPU experiments since the CPU characteristics of this second machine
do not affect performance while using the GPU.

\subsection{Synthetic ``Koller'' Data}\label{ssec:koller_data}

\begin{figure}[t]
\centering
\begin{subfigure}{.5\textwidth}
  \centering
  \includegraphics[width=0.9\linewidth]{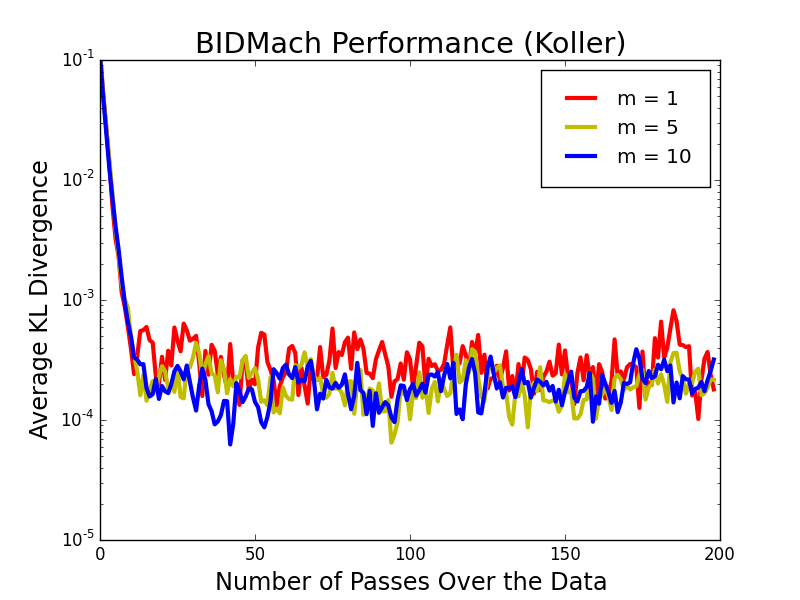}
  \caption{The $KL_{\rm avg}$ from BIDMach.}
  \label{fig:kl_bidmach}
\end{subfigure}%
\begin{subfigure}{.5\textwidth}
  \centering
  \includegraphics[width=0.9\linewidth]{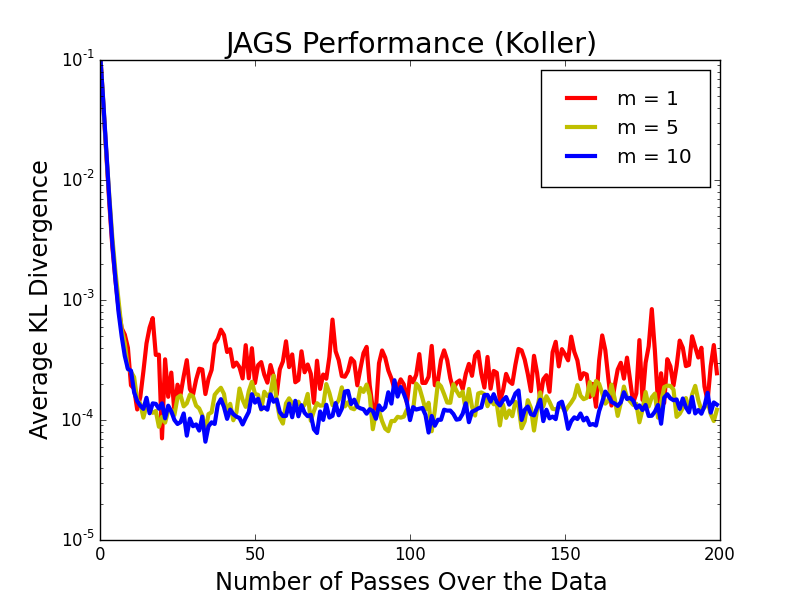}
  \caption{The $KL_{\rm avg}$ from JAGS.}
  \label{fig:kl_jags}
\end{subfigure}
\caption{Plots of our $KL_{\rm avg}$ curves as a function of the iteration, on Koller data (these
must be viewed in color). Both BIDMach and JAGS converge to the true CPTs quickly, but JAGS appears
to benefit more from the extra data replication with SAME parameters $m=5$ and $m=10$.}
\label{fig:first_set}
\end{figure}

\begin{table}[t]
\small
\caption{BIDMach (CPU) vs. JAGS Runtime on Koller Data}
\label{tab:bidmach_jags_koller}
\begin{center}
\begin{tabular}{ |c|c|c|c|c|c|c| } 
\hline
                         & $m=1$ & $m=2$ & $m=5$ & $m=10$ & $m=20$ \\
\hline \hline
BIDMach Total Time (sec) & 11.6  & 18.6  & 33.7  & 62.9   & 116.0  \\ 
JAGS Total Time (sec)    & 42.0  & 98.2  & 281.2 & 535.0  & 1037.6 \\
\hline
\end{tabular}
\end{center}
\end{table}

We first use synthetic data generated from a small Bayesian network to check correctness
and the use of SAME. The network has five variables: $X_0 = {\rm Intelligence}$, $X_1 =
{\rm Difficulty},$ $X_2 = {\rm SAT}$, $X_3 = {\rm Grade}$, and $X_4 = {\rm Letter}$. The directed
edges are $\mathcal{E} = \{(X_0, X_2), (X_0, X_3), (X_1,X_3), (X_3,X_4)\}$, where $(X_i,X_j)$ means
an arrow points from $X_i$ to $X_j$.  Variable $X_3$ is ternary, and all others are binary. This
network models a student taking a class, and considers ability metrics (Intelligence and SAT score),
the class difficulty, and the student's resulting grade, which subsequently affects the quality of a
letter of recommendation. This network, along with the true set of CPTs, is from Chapter 3
of~\citet{Koller2009}. Due to the name of the author, we call this the ``Koller'' data to
distinguish it from the MOOC data we use in~(\ref{ssec:mooc_data}).

To generate the data, we use the standard technique of forward sampling, where $X_i$ gets sampled
based on the true distributions from~\citet{Koller2009}, which depends on $X_i$'s parents (if any).
We repeat this to get 50,000 samples, then randomly hide 50\% of the data points. The goal is to
apply Gibbs sampling to estimate the CPTs that generated the data. Note that BIDMach can handle
millions of samples, but due to limitations of JAGS, we use only 50,000 for benchmarking.

To evaluate our Gibbs sampler, we compute the average KL Divergences of all the distributions in the
set of CPTs, denoted as $KL_{\rm avg}$.  For two distributions $p(x)$ and $q(x)$, the KL divergence
is $\sum_x p(x) \log(p(x)/q(x))$, summing over $x$ such that $q(x) > 0$.  In the Koller data, there
are eleven probability distributions that form the set of CPTs. For example, $X_4$ ``contributes''
three distributions: $\Pr(X_4 \mid X_3 = 0), \Pr(X_4 \mid X_3 = 1)$, and $\Pr(X_4 \mid X_3 = 2)$,
where $X_3$ (the parent) is fixed. For this network, we have $KL_{\rm avg} = \frac{1}{11}
\sum_{i=1}^{11} p_i(x) \log(p_i(x)/q_i(x))$, where $q_i$ is the distribution our sampler estimates
and $i$ is some arbitrary indexing notation. We do not use the KL Divergence of the full joint
distribution $P(X_1,X_2,X_3,X_4,X_5)$ since with high-dimensional data (e.g., the MOOC data
in~(\ref{ssec:mooc_data})), computing the full joint is intractable and we wish to facilitate
comparisons across different datasets.

Figures~\ref{fig:kl_bidmach} and~\ref{fig:kl_jags} plot the $KL_{\rm avg}$ metric for the student
data using BIDMach and JAGS, respectively with three SAME factors (note the log scale). The plots
indicate that the $KL_{\rm avg}$ for BIDMach and JAGS reach roughly the same values, with a slight
advantage to JAGS, though this is amplified because of the log scale. In practice, the difference
would be indistinguishable to humans. For instance, with $m=1$, the average difference between a
random number in the sampled CPT and its corresponding number in the true CPT for BIDMach and JAGS,
respectively, is 0.0039 and 0.0043. Thus, both BIDMach and JAGS will sample CPTs that are accurate
to two/three decimal places.

For BIDMach, we tuned the minibatch size to be 12,500. Using a smaller size means that, for a fixed
number of passes, we tend to get faster convergence, but this often comes with slower runtime per
iteration. In addition, we observed that increasing $m$ results in CPT estimates that more closely
match the true CPTs. The red curves (i.e., $m=1$) for Figures~\ref{fig:kl_bidmach}
and~\ref{fig:kl_jags} generally correspond to worse $KL_{\rm avg}$ than the respective yellow and
blue curves, with the difference more noticeable for JAGS. Also, going from $m=5$ to $m=10$ seems to
have a negligible benefit.

We further benchmark the speed of our CPU Gibbs sampler with JAGS on this data. For a fair
comparison, we keep our minibatch size to be 12,500 to match the runs from
Figure~\ref{fig:kl_bidmach}.  Table~\ref{tab:bidmach_jags_koller} shows the total runtime on the
data with different $m$ for 200 passes. We used total runtime because the JAGS API does not enable
us to separate the initialization time from the updating/sampling time. (JAGS spends a substantial
amount of time initializing since it needs to form a graph, but BIDMach uses matrices and
spends less than one second in initialization.) The results demonstrate that BIDMach is almost four
times faster than JAGS on the original data, and when the SAME parameter increases, the gap widens
(up to a factor of nine with $m=20$).

\subsection{Dynamic Learning Maps ``MOOC'' Data}\label{ssec:mooc_data}

\begin{figure}[t]
\centering
\begin{subfigure}{.5\textwidth}
  \centering
  \includegraphics[width=0.9\textwidth]{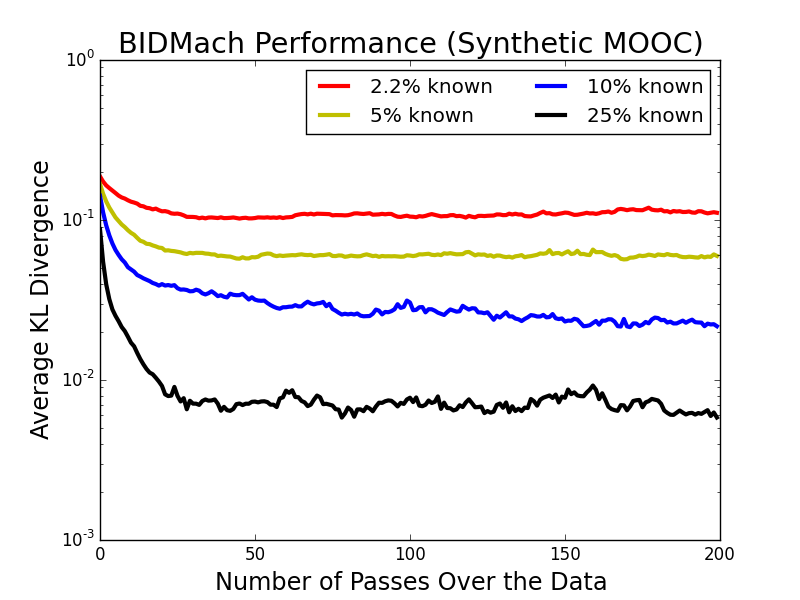}
  \caption{The $KL_{\rm avg}$ from BIDMach.}
  \label{fig:kl_bidmach_mooc}
\end{subfigure}%
\begin{subfigure}{.5\textwidth}
  \centering
  \includegraphics[width=0.9\textwidth]{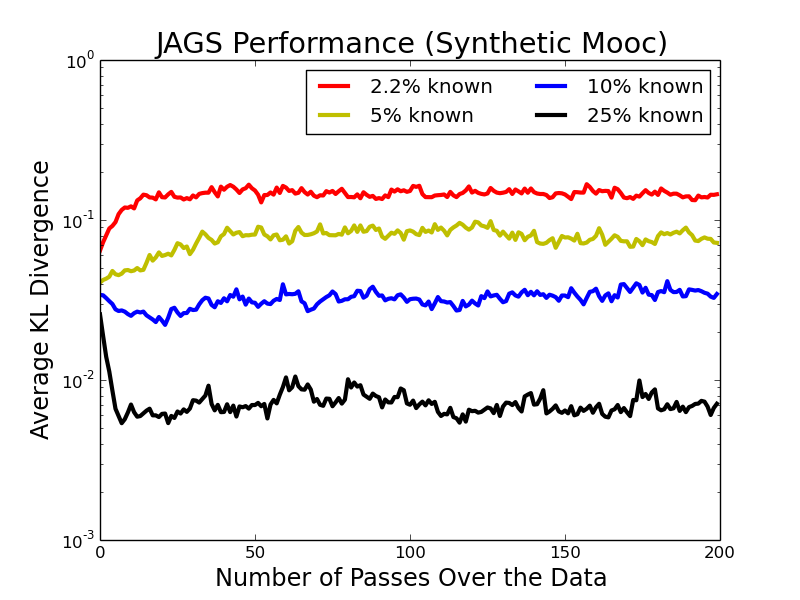}
  \caption{The $KL_{\rm avg}$ from JAGS.}
  \label{fig:kl_jags_mooc}
\end{subfigure}
\caption{Plots of our $KL_{\rm avg}$ curves as a function of the iteration, on Synthetic MOOC Data
with different sparsity levels (these must be viewed in color).}
\label{fig:second_set}
\end{figure}

\begin{figure}[t]
\centering
\begin{subfigure}{.5\textwidth}
  \centering
  \includegraphics[width=0.9\textwidth]{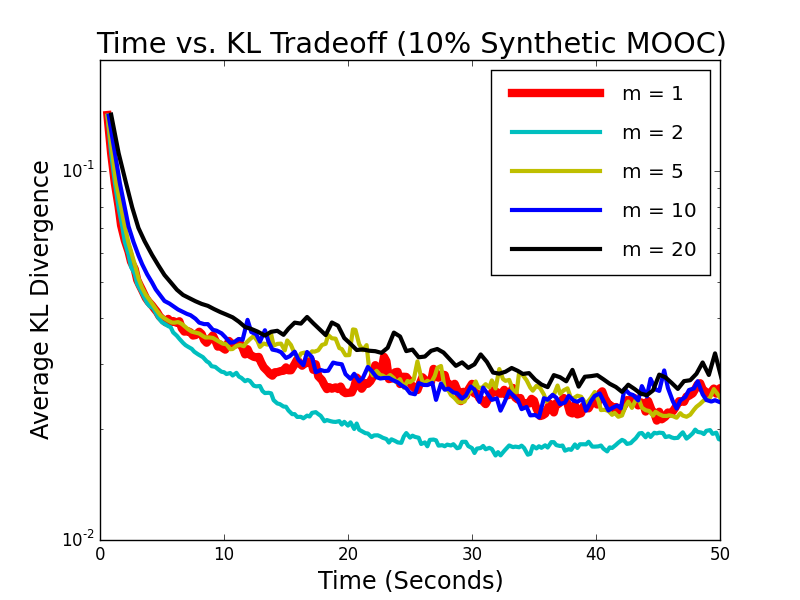}
  \caption{The $KL_{\rm avg}$ from BIDMach.}
  \label{fig:mooc_kl}
\end{subfigure}%
\begin{subfigure}{.5\textwidth}
  \centering
  \includegraphics[width=0.9\textwidth]{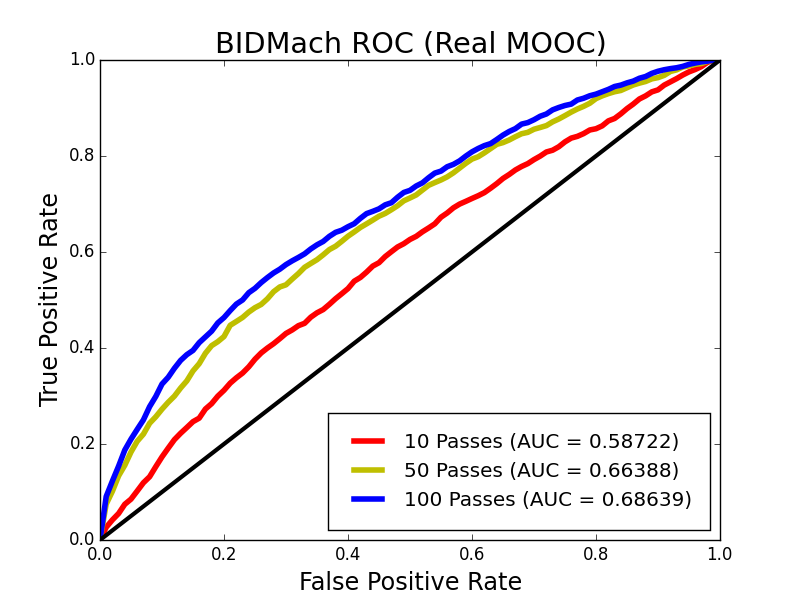}
  \caption{MOOC data ROC curve.}
  \label{fig:mooc_accuracy}
\end{subfigure}
\caption{Some more results ($KL_{\rm avg}$ and ROCs) of our sampler on Synthetic/Real MOOC data.}
\label{fig:third_set}
\end{figure}

We now benchmark our code on a Bayesian network with a nation-wide examination dataset from the
Dynamic Learning Maps (DLM) project\footnote{URL: \url{http://dynamiclearningmaps.org/}. There is no
official paper reference for this data.}. The data contains the assessment (correct or not) of
student responses to questions from the DLM Alternate Assessment System. After preprocessing, there
are 4367 students and 319 questions. Each of the questions is derived from a set of 15 basic
concepts. We encode questions and concepts as variables and describe their relationships with a
Bayesian network from the DLM data. (The 15 concepts are latent variables across all students.) Each
question is a variable with a very high missing value rate.  Our corresponding data matrix, which
has dimensions $(334\times 4367)$, has a 2.2\% density level. The inference task is to learn the
CPTs of the Bayesian network from this extremely sparse data. All variables are binary.

We evaluate our sampler using two methods. The first involves running it on the original data until
convergence. Then we use the resulting estimated set of CPTs and sample from that via forward
sampling to generate ``Synthetic MOOC'' data, also of dimension $(334 \times 4367)$. We create
several versions of this data, each with a different fraction of missing data. Then we re-run
BIDMach on these and evaluate the CPTs using the same $KL_{\rm avg}$ metric
from~(\ref{ssec:koller_data}). Computing $KL_{\rm avg}$ involves averaging over 682 distributions
based on the Bayesian network, which has relatively few edges (the most amount of parents any node
has is three).

Figures~\ref{fig:kl_bidmach_mooc} and~\ref{fig:kl_jags_mooc} plot the $KL_{\rm avg}$ for BIDMach and
JAGS, respectively on four different levels of sparsity (note again the log scale). As expected,
BIDMach and JAGS converge to roughly the same $KL_{\rm avg}$. Furthermore, denser data allows
the Gibbs sampler to get closer to the true CPTs. Note that BIDMach and JAGS start at different
$KL_{\rm avg}$ values due to different initializations (BIDMach initializes randomly, while JAGS
tends to start with ``even'' $(0.5,0.5)$ distributions), and the JAGS curve increases for extremely
sparse data because the initialization point happens to be better than to what it converges.

We also analyze the tradeoffs involved in increasing the SAME factor $m$. While increasing $m$
intuitively (and theoretically) should lead to faster convergence to higher quality parameter
estimates, it is possible that increasing $m$ too much will result in getting stuck in a local
maxima since this is a highly noncovex problem. Figure~\ref{fig:mooc_kl} plots\footnote{The baseline
$m=1$ curve is thicker for readability.} $KL_{\rm avg}$ as a function of total time elapsed. The
$m=2$ curve is clearly better as for a fixed time $t$, it has lower $KL_{\rm avg}$ than $m=1$,
despite how the $m=1$ curve will have had more full passes over the data. Interestingly enough,
$m=5$ results in similar results, and the $m=20$ curve indicates that higher $m$ will not be
beneficial.  We ran the experiment for Figure~\ref{fig:mooc_kl} for 300 seconds and the longer-term
trend was that the $KL_{\rm avg}$ values tended to remain at their 50-second values.

In addition to measuring distance between predicted and actual parameters, we also evaluate the
sampler using its accuracy at predicting missing labels.  We randomly split the original data into a
training and testing batch so that 80\% of the known data is in training. The training set is used
to update the CPTs. During the testing phase, we sample $n$ times using the current CPTs without
updating them. For each (correct) test state, this yields an unbiased estimate of the correct state
probability (note that this is a binary classification problem).




The training and testing data have 61.9\% and 61.3\% of known data points in the positive class,
respectively.  Due to this imbalance, we do not use raw accuracy, but instead use Receiver Operating
Characteristic (ROC) curves to estimate the predictor's effectiveness.
Figure~\ref{fig:mooc_accuracy} shows the ROC curves obtained after 10, 50, and 100 passes through
the data, with increasing AUCs of 0.58722, 0.66388, and 0.68639, respectively,  indicating that our
sampler is effective on this extremely sparse data.


\section{Speed and Scalability of Our Sampler on Large Data}\label{sec:scaling_large_data}

%
%
%
%
%
%
%
%
\begin{table}[t]
\small
\caption{BIDMach (CPU) vs. JAGS Runtime on (Replicated) Real MOOC Data}
\label{tab:bidmach_jags_realmooc}
\begin{center}
\begin{tabular}{ |c|c|c|c|c|c|c| } 
\hline
                             & 1x    & 2x     & 5x     & 10x     & 20x     & 40x    \\
\hline \hline
GPU BIDMach Total Time (sec) & 11.2  & 21.2   & 48.1   & 92.9    & 183.8   & 365.9  \\ 
CPU BIDMach Total Time (sec) & 39.5  & 76.1   & 187.3  & 359.6   & 701.4   & 1437.4 \\ 
JAGS Total Time (sec)        & 975.2 & 2749.0 & 5830.0 & 18815.0 & 34309.0 & OOM    \\
\hline
\end{tabular}
\end{center}
\end{table}

\begin{table}[t]
\small
\caption{BIDMach (GPU) Runtime Per Iteration and GigaFlops on Large Data}
\label{tab:tradeoff}
\begin{center}
\begin{tabular}{ |c|c|c|c|c|c| } 
\hline
               & $m=1$ & $m=2$ & $m=5$ & $m=10$ \\
\hline \hline
GigaFlops (Koller, 1M)               & 2.38  & 4.07  &  6.98  &  9.08  \\ 
Time (sec) / Iteration (Koller, 1M)  & 0.307 & 0.359 &  0.523 &  0.804 \\
\hline
GigaFlops (30x MOOC)              & 2.69  & 4.46  & 7.83  & 10.37   \\ 
Time (sec) / Iteration (30x MOOC) & 2.440 & 2.940 & 4.181 & 6.313   \\
\hline
\end{tabular}
\end{center}
\end{table}

We now discuss the extent to which BIDMach (and JAGS) can scale up to larger datasets in order to
identify the limits of current software. To obtain arbitrarily large datasets, we replicated the
MOOC data (from~(\ref{ssec:mooc_data})) horizontally: we took the data matrix and made $n$ copies to
get a $(334 \times 4367n)$-dimensional data matrix. In Table~\ref{tab:bidmach_jags_realmooc}, we
report the total runtime for 200 iterations for BIDMach and JAGS on a variety of replicated data,
from the original data (1x) to data replicated 40 times (40x). For our sampler, we used the same
tunable settings (e.g, minibatch size) as we did in Figure~\ref{fig:second_set}, to keep our
benchmarks consistent.

The results indicate that BIDMach holds a clear speed advantage over JAGS (for both the CPU and
GPU), though one must interpret the results carefully. From Figure~\ref{fig:second_set}, we see
that, while BIDMach and JAGS both eventually converge to similar $KL_{\rm avg}$ values, our sampler
takes roughly three times as many iterations to reach convergence (this is most easily observed on
the 25\% data).

Therefore, to determine BIDMach's speed advantages, one would need to divide the JAGS time by
three (roughly). Despite this caveat, our sampler is still substantially faster. With the
original data, BIDMach's advantage is a factor of $(975.2/3)/11.2 \approx 29.0$ (with the GPU), and
the gap widens with more data (for 20x, we have an advantage of $(34309.0/3)/183.8\approx 66.2$).
In terms of memory usage, BIDMach also outperforms JAGS. BIDMach structures data in matrices,
whereas JAGS internally forms a graph.  For the 40x data, JAGS ran out of memory on our 64 GB CPU
RAM machine, whereas BIDMach used up only 11.4\% of the CPU RAM.

It is also worthy to investigate the impact of the SAME parameter $m$, in terms of runtime \emph{and
throughput}, which is measured in GigaFlops (gflops), a billion floating point operations per
second. As the SAME parameter increases, it increases both the throughput and runtime by increasing
the number of data points in our computations. The results from~\citep{SAME2015} suggest that
increasing $m$ for small values will increase throughput while not costing too much in runtime.  As
one increases $m$ beyond a certain data-dependent value, then SAME ``saturates'' the algorithm and
results in stagnant throughput while significantly increasing runtime.  Figure~\ref{fig:mooc_kl}
also raises the question as to whether one really needs a large $m$ in the first place.

Table~\ref{tab:tradeoff} reports the performance of BIDMach's GPU version (only) on two expanded
datasets with varying $m$. One is the Koller data from~(\ref{ssec:koller_data}), with 50\% of the
data known versus missing, but with \emph{one million} cases. The other is 30x replicated MOOC data.
Table~\ref{tab:tradeoff} reports that BIDMach gets a healthy amount of throughput from the expanded
data (especially the 30x MOOC data), and that increasing $m$ further is a logical possibility.

Finally, we point out that the runtimes in all tables in this paper are listed are in seconds, and
even with large $m$ and a large replication factor, we have barely scratched the limit of our
sampler. To test the limit of our GPU sampler, we ran it on 300x replicated MOOC data with $m=1$. Our sampler
completed 200 iterations in 1726.1 seconds with a gflops of 7.59.  In fact, the primary limiting
factor of our sampler is not the size of the data itself, but the memory of the GPU; the 300x data
used up almost all of the 12GB of GPU RAM. As we get better and faster GPUs, the performance of our
sampler will improve and we will be able to run it on larger and larger datasets.

\section{Conclusions}\label{sec:conclusions}

We conclude that our Gibbs sampler is much faster than the state of
the art (JAGS) and can be applied to data with hundreds of discrete
variables and millions of instances. We also suggest that SAME can be
beneficial for Gibbs sampling, and that SAME Gibbs sampling should be
the go-to method for researchers who wish to perform inference on
(discrete) Bayesian networks. Future work will explore the application
of our sampler to a wider class of real-world datasets.

\subsubsection*{Acknowledgments}

We thank Yang Gao, Biye Jiang, and Huasha Zhao for helpful discussions.

\bibliography{iclr2016_conference}
\bibliographystyle{iclr2016_conference}

\end{document}